\documentclass[11pt]{article}

\usepackage[final]{acl}

\usepackage{times}
\usepackage{latexsym}

\usepackage[T1]{fontenc}

\usepackage[utf8]{inputenc}

\usepackage{microtype}

\usepackage{inconsolata}

\usepackage{graphicx}
\usepackage{makecell}
\usepackage{times}
\usepackage{latexsym}
\usepackage{multirow} 
\usepackage{hyperref}
\usepackage{booktabs} 
\usepackage{array}
\usepackage{makecell}
\usepackage{graphicx}
\usepackage[inline]{enumitem}
\usepackage{amssymb} 
\usepackage[table]{xcolor}
\usepackage{pgfplots}
\pgfplotsset{compat=1.18}
\usepackage{pgfplotstable}
\usepackage{caption}
\usepackage{float}
\usepackage{pgfmath}
\usepackage{comment}
\usepackage{lipsum} 
\usepackage{fancyvrb} 
\renewcommand{\arraystretch}{1.2}

\newcolumntype{L}[1]{>{\raggedright\arraybackslash}p{#1}}
\newcolumntype{C}[1]{>{\centering\arraybackslash}p{#1}}

\usepackage{amsmath}

\newcolumntype{L}[1]{>{\raggedright\arraybackslash}m{#1}} 
\renewcommand{\arraystretch}{1.4}

\newcommand{\gio}[1]{{\color{black}#1}}
\newcommand{\eleni}[1]{#1}
\newcommand{\elenin}[1]{#1}

\usepackage[T1]{fontenc}

\usepackage[utf8]{inputenc}

\usepackage{microtype}

\usepackage{inconsolata}

\title{Beyond Logical Forms: LLM-Extracted Patterns for Fallacy Classification}

\author{Eleni Papadopulos$^{1,2}$, Firoj Alam$^3$ Giovanni Da San Martino$^2$\\
  $^1$Politecnico di Torino, Italy,
  $^2$Università di Padova, Italy \\
  $^3$Qatar Computing Research Institute, Qatar\\ 
  \texttt{eleni.papadopulos@polito.it, giovanni.dasanmartino@unipd.it}\\
  \texttt{fialam@hbku.edu.qa}\\
}

\begin{document}
\maketitle
\begin{abstract}

In today's fast-paced information era, logical fallacies, defined as defective patterns of reasoning, inevitably contribute to the growth of information disorder. However, often fallacies appear in nuanced forms that complicate automated classification. In this study, we investigate whether \eleni{merging abstract logical structures with context-level linguistic cues} proves beneficial for fallacy classification, developing a framework that inductively extracts such patterns from fallacious examples and their explanations using Large Language Models (LLMs). 
We evaluate the impact of these patterns across different LLMs and experimental zero- and one-shot configurations, showing statistically significant improvements over zero-shot baselines and outperforming competing approaches.  
Cross-dataset experiments validate generalization, establishing data-driven pattern extraction as an effective method for generating logical representations. 

\end{abstract}

\section{Introduction\label{sec:intro}}

A logical fallacy is a common thinking error, especially one apt to mislead \citep{gensler2010z}. These arguments often appear rational and logically coherent on the surface, but deeper analysis reveals they are not \citep{Copi1953-COPITL-9}. Fallacies are traditionally classified into formal and informal types: \textbf{\textit{formal fallacies}} violate the rules of 
logical structure regardless of content, while \textbf{\textit{informal fallacies}} are patterns of mistakes that are made in the everyday uses of language and are related to contextual meaning \cite{hamblin1970fallacies, bacon1999logic}.

To evaluate the quality of an argument, it is helpful to reconstruct it into what is known as logical form, the structure that emerges when the specific content of a statement is replaced by variables \citep{Johnson1977-JOHLS}. For example, the argument \textit{If it rains, then the ground will be wet. It is raining. Therefore, the ground is wet} has the logical form \textit{If P, then Q. P. Therefore, Q}. 
Building on this formalization framework,
\citet{jin2022logicalfallacydetection} developed a structure-aware model for fallacy detection on the \textsc{Logic} dataset that compares arguments' and fallacies' logical forms. 
\gio{However, in their approach a single logical form is assigned to each fallacy, which might fail to capture the full spectrum of ways a fallacy can manifest in natural discourse. 
Another challenge is related to
informal fallacies, where reasoning is often more nuanced and context-dependent than abstract representations suggest.
} 

\elenin{These limitations motivate the need to go beyond purely abstract representations, incorporating linguistic elements, such as lexical markers or rhetorical devices, to provide a more comprehensive characterization of how fallacies manifest in natural language. We argue that LLMs can inductively extract such representations from fallacious examples, capturing both the logical structure and linguistic cues that reveal the underlying mechanisms of deception. \eleni{Unlike prior work that formalized fallacy logic through hand-crafted templates \citep{robbani-etal-2024-flee}, our approach is data-driven and not restricted to a limited number of fallacies.} Hereafter, we refer to these extracted structures collectively as \textit{structural patterns}. Our goal is to investigate whether context-aware structural information is valuable for automated fallacy detection.}

While existing supervised approaches require 
heavy computational resources for fine-tuning \citep{vijayaraghavan-vosoughi-2022-tweetspin, lei2024boostinglogicalfallacyreasoning, sourati2023casebasedreasoninglanguagemodels, sourati2023robustexplainableidentificationlogical, alhindi-etal-2024-large}, to our knowledge, no prior work has explored fallacy classification from a structural perspective 
without any additional fine-tuning. Although we use labeled data for pattern extraction \eleni{(resulting in a weakly supervised approach),} our 
framework avoids fine-tuning costs and produces generalizable patterns that allow classification through prompting alone, enabling comparison with unsupervised methods.

We evaluate multiple prompting configurations to determine which 
components enhance performance and examine the impact of demonstrations on detection capabilities. Our approach, incorporating generated patterns, achieves noteworthy results among unsupervised methods on the dataset \textsc{Logic}.
Finally, to validate the robustness and transferability of our patterns, we assess their performance across two further datasets spanning diverse domains and argumentative styles.

In summary, our contributions are threefold:
\begin{itemize} 
    \item We leverage LLMs to automatically extract patterns from fallacious examples and their explanations, which are then employed in inference-only classification.
    \item We evaluate different LLMs with \gio{various} prompt designs outperforming competing approaches on the \textsc{LOGIC} dataset. 
    \item We validate generalizability by testing our patterns on two different datasets with different domains and structures.
\end{itemize}

\section{Related Work}

Recent advances in fallacy detection have increasingly turned to LLMs, though few studies have relied exclusively on prompting-based techniques. 
Several works have employed fallacy detection to probe LLMs' logical reasoning abilities \citep{llm4lfd, hong2024closerlookselfverificationabilities, li2024reasonfallacyenhancinglarge, xu2025socratessmartypantstestinglogic}.  
Among these, \citet{hong2024closerlookselfverificationabilities} investigated self-verification capabilities and showed
that LLMs face more challenges with structure-based 
(formal) fallacies with respect to content-based (informal) ones, and that fallacy definitions provide minimal improvements. 
\citet{xu2025socratessmartypantstestinglogic} has shown that reasoning models have better performances with respect to non-reasoning ones for fallacy classification. 
Among studies relying exclusively on prompting techniques, \citet{pan2024llmsgoodzeroshotfallacy} designed single-round and multi-round prompting schemes for zero-shot detection, while \citet{jeong2025largelanguagemodelsbetter} introduced contextual prompting incorporating counterarguments, explanations, and goals with confidence-based ranking, showing that explanations particularly enhance performance. \citet{lim2024evaluationllmidentifyinglogical} assessed detection abilities on the \textsc{Logic} dataset using few-shot prompting, though their different taxonomy limits direct comparison with our work. Other research has examined the logical structure of argumentation. Most notably, \citet{jin2022logicalfallacydetection} developed a structure-aware model based on Electra that distills arguments into logical forms and compares them against fallacy patterns sourced from \href{https://www.logicallyfallacious.com/}{logicallyfallacious.com}. \eleni{Another prominent framework in this field is Walton's theory of argumentation \citep{alma9924387453302466}, consisting in about 60 templates that capture common argument types, each associated with a set of critical questions to evaluate their validity. These schemes have been  adopted in computational approaches for disinformation \citep{gutierrez2024detecting}, misinformation \citep{ruizdolz2025explainableframeworkmisinformationidentification} and fake news \citep{wangcabrio} detection. Regarding fallacies specifically, \citet{ruiz-dolz-lawrence-2023-detecting} introduced a dataset of sentences grounded in Walton's argumentation schemes, labeling them as fallacies when the associated critical questions could not be successfully answered. However, Walton's schemes do not hold a one-to-one correspondence with fallacies and cover only a limited number of them, limiting their direct applicability to fallacy classification. Of particular relevance is the work of \citet{robbani-etal-2024-flee}, who re-designed four of \citet{alma9924387453302466} and \citet{reisert-etal-2018-feasible}'s schemes with the goal of expliciting fallacies' implicit logic, introducing formal logical schemas with explicit variables and relationships. While this represents a meaningful step toward structural formalization, their patterns are manually designed and leave a large portion of fallacy types unrepresented. 
Our patterns, by contrast, are extracted automatically from data, allowing to adapt to intra-class variation.}

\section{Datasets\label{sec:datasets}}

The \textsc{Logic} dataset is a collection of 2,449 examples across 13 fallacy types~\citep{jin2022logicalfallacydetection}. Instances are sourced from educational platforms about fallacies such as Quizziz and study.com. The dataset consists of brief dialogues and short statements.
Given the educational intent behind these examples, sentences tend to have relatively straightforward syntactic structures, making the dataset particularly well-suited for the extraction of the patterns. 

Although it contains 13 distinct classes, a thorough analysis revealed that some of the classes actually contain instances of different fallacies, that were grouped together. For instance, the class \textit{Hasty Generalization} contains examples of actual \textit{Hasty Generalization} as well as \textit{Slippery Slope} (Table~\ref{tab:class-fallacies}). 
While these grouped fallacies share common logical flaws and thus belong to the same conceptual group, they manifest through different structural patterns. 

\begin{table}[t]
\renewcommand{\arraystretch}{1.0} 
\centering
\scriptsize
\resizebox{\columnwidth}{!}{%
\begin{tabular}{ll}
\toprule
\emph{\textbf{Class}} & \emph{\textbf{Fallacies included}} \\
\midrule
\textbf{Intentional Fallacy}  & \begin{tabular}[t]{@{}l@{}}Intentional Fallacy\\ Shifting the Burden of Proof\\ Moving the Goalposts\\ No True Scotsman\end{tabular} \\ \midrule
\textbf{False Cause}          & \begin{tabular}[t]{@{}l@{}}Post Hoc\\ False Cause\end{tabular}                                                                      \\ \midrule
\textbf{Hasty Generalization} & \begin{tabular}[t]{@{}l@{}}Hasty Generalization\\ Slippery Slope\end{tabular}                                                       \\ \bottomrule
\end{tabular}%
}
\caption{Examples of classes in \textsc{Logic} dataset containing instances of different fallacy types. While 
coherent, these groupings comprise fallacies with distinct structural patterns.  A detailed breakdown of all classes' subtypes is provided in \mbox{Appendix D} of \citet{jin2022logicalfallacydetection}.}

\label{tab:class-fallacies}
\end{table}
%
We experiment on two further datasets: 
\textsc{Reddit} \citep{sahai-etal-2021-breaking}, consisting of fallacious comments extracted from subreddits covering different topics and \textsc{ElecDebate60to16} (hereafter \textsc{ElecDebate}) \citep{goffredo-etal-2023-argument}, a collection of televised debates of the presidential election campaigns in the U.S. from 1960 to 2016. Some fallacy classes contain sub-categories. In Table~\ref{tab:fallacy-datasets}, we report a summary of the dataset, and a description of each taxonomy is provided in Appendix~\ref{sec:appendix-datasets}.



\begin{table}[ht]
\centering
\setlength{\tabcolsep}{2pt} 
\scalebox{0.73}{
\begin{tabular}{lccll}
\toprule
\textbf{Data} & \textbf{Dataset split} & \textbf{\# Classes} & \textbf{Genre} & \textbf{Domain} \\
\midrule
\textsc{Logic}      & 1807/299/299 & 13         & Dialogue      & Education \\
\textsc{Reddit} & 588/148/105   & 8\textsuperscript{\ddag}  & Comments     & General \\
\textsc{ElecDebate}   & 1120/200/187  & 6  & Dialogue  & Politics \\
\bottomrule
\end{tabular}
}
\caption{
Statistics of the three datasets. \ddag\ indicates that the \textit{No Fallacy} class is included.
\label{tab:fallacy-datasets}}
\end{table}

\section{Pattern Generation}
\label{sec:pattern-generation}
Natural arguments appear in several  forms. Such variability manifests itself in \textsc{Logic} dataset as well as many others \cite{habernal-etal-2018-name, da-san-martino-etal-2019-fine}.
For this reason, we address our research question by modeling patterns inductively from the training set of \textsc{Logic}. 
The choice of the 
dataset for pattern extraction is critical. It provides the required combination of structural clarity and fallacy diversity through its multiple sub-types per class. These properties make it especially suited for our purpose. The clean argumentative structure allows to formalize clear logical patterns while capturing intra-class variations.

Our pattern generation procedure features two steps:

\textbf{Step 1: Explanation Generation} 
Explanations have been shown to be instrumental in identifying and discrediting fallacious reasoning, as they make the logical structure of arguments explicit and open to scrutiny \citep{Storer_1949}. 
Furthermore, \citet{jeong2025largelanguagemodelsbetter} has demonstrated that providing explanations constitutes valuable contextual information in zero-shot settings. 
We expected explanations to facilitate pattern extraction by breaking down the reasoning process and revealing shared reasoning flaws, particularly useful for informal fallacies.

Given a sentence from the training set and its fallacy label, we used \texttt{llama-3.3-70B-Instruct}~\cite{dubey2024llama} to generate an explanation that justifies why that sentence contains the specified fallacy.

\textbf{Step 2: Pattern Extraction} For each fallacy class, we used OpenAI's reasoning model \texttt{o4-mini} \citep{o4} to extract patterns from all the sentences of that class and their explanations, requiring the model to preserve function words such as prepositions or adverbs and to abstract away from content words by 
\gio{replacing them with} placeholders while keeping the original reasoning form. 
Additionally, summaries were extracted to derive new fallacy definitions.

We opted for \texttt{llama-3.3-70B-Instruct} for explanation generation as it provided high-quality explanations while remaining cost-effective for large-scale text generation. For pattern extraction, we employed \texttt{o4-mini} given its reasoning capabilities. \elenin{The prompts used in our experiments are reported in the Github repository.\footnote{\label{fn1}\url{https://github.com/elenipapadopulos/fallacy-patterns}}}

In the initial phase of our research, we aimed to cover two distinct logical aspects from our arguments and explanations, specific to formal and informal fallacies, respectively:
\begin{itemize}
    \item arguments' \textbf{logical 
    \elenin{structure}} inspired by formal logic theory;
    \item recurring \textbf{\elenin{lexical} schemes} that frequently appear in both sentences and explanations, capturing specific information about the reasoning behind the fallacy as well as frequent syntactic particles, phrases, and examples that convey the fallacious intent. 
\end{itemize}
Our patterns incorporate both of these aspects
, as Table~\ref{tab:int-fal} shows. These \textsc{logic}-based patterns combine reasoning structure (variables X, Y, Z) with concrete linguistic features (specific phrases, loaded terms, rhetorical devices), occasionally retaining some definitions. The full list of patterns is available in the repository.\footref{fn1} 

The process resulted in approximately 3-6 patterns per fallacy class. Final patterns were obtained after providing different subsets to the model and selecting the best performing one on the validation set, in the attempt to retain only useful information and avoid redundancy. 
In section~\ref{sec:datasets} we discussed how one class in the datasets could correspond to multiple fallacies. 
Although in some cases, e.g. a pattern for \textit{Tu quoque} (a fallacy which is part of the class  \textit{Ad Hominem} in \textsc{Logic}), is correctly generated and selected, sometimes fails to select patterns when multiple fallacies are grouped under the same class label.
This is expected because we include the fallacy class name in the prompt, which likely biases the model toward patterns that match its internal knowledge of that particular class name. To ensure a broader coverage of fallacies listed in Table~\ref{tab:class-fallacies}, we manually isolated instances of frequent and undetected fallacies (such as \textit{Shifting the Burden of Proof}) and repeated the procedure.

\begin{table}[!h]
\renewcommand{\arraystretch}{1.05}
\centering
\small
\begin{tabular}{@{}p{\columnwidth}@{}}
\toprule
\centering\emph{\textbf{Intentional Fallacy Patterns}} \tabularnewline
\midrule
\vspace{-6pt}
\begin{enumerate}[nosep,leftmargin=*,itemsep=1pt]
    \item The argument assumes that because X (e.g., someone’s intention, belief, or lack of counter-evidence), therefore Y is true.
    \item Asserting P is true because it has not been disproven.
    \item Because the creator intended [interpretation], the work should be understood as [interpretation].
    \item Questions framed to presuppose guilt or a specific intention (e.g., ``Have you stopped X?''), thus assuming what is to be proven.
    \item If A does not have trait X, and X is allegedly typical of group G, then A is not a member of G.
\end{enumerate}
\\[2pt]
\midrule
\centering\emph{\textbf{Red Herring Patterns}} \tabularnewline
\midrule
\vspace{-6pt}
\begin{enumerate}[nosep,leftmargin=*,itemsep=1pt]
    \item Instead of addressing [original issue], the argument shifts focus to [irrelevant topic], which distracts from the main discussion.
    \item The argument attempts to justify, explain, or defend by referencing [irrelevant detail], ignoring the original issue of [main topic].
    \item A shift from the initial question or problem to a secondary topic that does not logically follow, e.g., ``You asked about X, but I will tell you about Y.''
\end{enumerate}
\\[2pt]
\bottomrule
\end{tabular}
\caption{Patterns for \textit{Intentional Fallacy} and \textit{Red Herring}. For \textit{Intentional Fallacy}, patterns (\#4) and (\#5) illustrate lexical schemes and logical forms, respectively, that encode intent and structure.}
\label{tab:int-fal}
\end{table}

\section{Experiments}
\label{sec:experiments}

This section describes our experiments for fallacy classification, including our patterns extracted by the procedure introduced in Section~\ref{sec:pattern-generation} and several competing prompting strategies. Additional experiments are reported in Appendix~\ref{sec:additiona-experiments}.
We used the following LLMs for our experiments: \texttt{gpt-4o}, \texttt{o4-mini}, \texttt{gpt-4.1-mini}, \texttt{LLama-3.3-70B}, \texttt{deepseek-r1} and \texttt{Gemma-3-27B-it} for a total cost of 75 USD. Our intent was to test LLMs from different providers and with different sizes and to compare reasoning and non-reasoning models.

\subsection{Prompt Design}

\paragraph{Baselines.} 
We compared our approach against several baselines that vary in the type and amount of information provided to the model. The simplest baseline (\textbf{\textsc{zero-shot}}) provides only the list of fallacy names in the dataset as a reference, establishing a minimal information condition. 
Our second baseline incorporates fallacy definitions to provide more comprehensive background knowledge (\textbf{\textsc{def}}). These definitions were initially sourced from \citet{lei2024boostinglogicalfallacyreasoning} and subsequently refined based on our analysis to ensure clarity and consistency. Finally, we tested a baseline using standard logical forms, following the approach of \citet{jin2022logicalfallacydetection} and sourcing these forms from \href{https://logicallyfallacious.com/}{logicallyfallacious.com}. This final baseline (\textbf{\textsc{logical forms}}) allows us to assess the effectiveness of expert-made logical representations compared to our generated pattern-based approach.

\paragraph{LLM-derived Patterns and Definitions.} 
Beyond generating structural patterns, we leveraged the explanations from \mbox{Section~\ref{sec:pattern-generation}} to automatically create new fallacy definitions based on \textsc{Logic} training samples. 
We then replicated experiment \textbf{\textsc{def}} with these new definitions (\textbf{\textsc{new def}}). 
We also exploited the patterns extracted by adding them to the prompt 
(\textbf{\textsc{patterns}}) 
and by implementing a two-step approach where we first ask the LLM to identify the pattern and then to output the corresponding fallacy  (\textbf{\textsc{pattern matching}}). 

\paragraph{One-shot Prompting.}
We further investigated the impact of providing examples to the model through several experimental configurations \citep{NEURIPS2020_1457c0d6}, with one-shot prompting proving most effective. Initially, we tested a static approach where one example per fallacy was randomly selected and shown to all test sentences (\textbf{\textsc{one-shot}}), establishing a baseline for example-based learning. To enhance this approach, we augmented the same examples with manually crafted explanations following our previously established definitions as guidelines (\textbf{\textsc{one-shot + exp}}). We sampled 5 different example sets and performance across all configurations was assessed over 5 runs to ensure statistical reliability. 

More sophisticated was our dynamic one-shot prompting approach (\textbf{\textsc{dynamic one-shot}}), which computes embeddings for both training and test sentences to retrieve, for each test sentence, the most similar example per class in the training set. We used \texttt{sentence-transformers/all-MiniLM-L6-v2}\footnote{\url{https://huggingface.co/sentence-transformers/all-MiniLM-L6-v2}} model and \texttt{cross-encoder/stsb-roberta-base}\footnote{\url{https://huggingface.co/cross-encoder/stsb-roberta-base}} cross-encoder from SentenceTransformers \citep{reimers-2019-sentence-bert} to compute embeddings and employed cosine similarity to evaluate similarity. We included the previously generated explanations of examples in the prompt as well (\textbf{\textsc{dynamic + exp}}).

Furthermore, we explored structure-focused similarity. Since \citet{jin2022logicalfallacydetection} released a version of \textsc{Logic} with masked arguments (with content words replaced by placeholders), we conducted the same similarity-based procedure using these masked sentences (see an example in Table~\ref{tab:masked-logic}) in the attempt to force the embedding model to focus on structural rather than lexical similarities. For this configuration (\textbf{\textsc{syntax-based dynamic one-shot}}), we used \texttt{sentence-transformers/all-MiniLM-L6-v2} from SentenceTransformers alongside a syntax-augmented version of RoBERTa-large extracted from \citet{sachan2021syntaxtreeshelppretrained} (see Appendix~\ref{sec:syntax-augm}).

Finally, we incorporated the generated patterns into our dynamically retrieved examples and their explanations (\textbf{\textsc{dynamic + exp + patterns}}).

\begin{table}[!h]
\renewcommand{\arraystretch}{1.05}
\centering
\small
\begin{tabular}{@{}m{0.18\columnwidth}m{0.74\columnwidth}@{}}
\toprule
\textbf{Original argument} &
Every time I wear this necklace, I pass my exams. Therefore, wearing this necklace causes me to pass my exams. \\
\midrule
\textbf{Masked argument} &
Every time MSK<0> MSK<2>, MSK<0> MSK<4>. Therefore, MSK<2> causes MSK<0> to MSK<4>. \\
\bottomrule
\end{tabular}
\caption{Example of a masked argument in \textsc{Logic}. The distillation algorithm is explained in \citet{jin2022logicalfallacydetection}. The masked version of the dataset was publicly released by the authors and was not created by us.}
\label{tab:masked-logic}
\end{table}


\paragraph{Multi-step Classification.} 

An alternative approach 
decomposes the classification task into three sequential steps within a single model call (\textbf{\textsc{multistep}}) using chain-of-thought prompting \citep{wei2023chainofthoughtpromptingelicitsreasoning}. In the first step, the model is required to generate a structural pattern from the argument according to predefined structural rules. 
Subsequently, the model should match it to one of the patterns and, as a result, classify the argument.

\subsection{Results and discussion}

\begin{table*}[!t]
\renewcommand{\arraystretch}{1}
\centering
\setlength{\tabcolsep}{3pt} 
\scalebox{0.75}{
\begin{tabular}{l*{6}{cc}}
\toprule
\textbf{Method} 
& \multicolumn{2}{c}{\textbf{o4-mini}} 
& \multicolumn{2}{c}{\textbf{gpt-4o}} 
& \multicolumn{2}{c}{\textbf{deepseek-r1}} 
& \multicolumn{2}{c}{\textbf{gpt-4.1-mini}} 
& \multicolumn{2}{c}{\textbf{llama-3.3-70B}} 
& \multicolumn{2}{c}{\textbf{gemma-3-27b-it}} \\
\cmidrule(lr){2-3} \cmidrule(lr){4-5} \cmidrule(lr){6-7} \cmidrule(lr){8-9} \cmidrule(lr){10-11} \cmidrule(lr){12-13}
& \textbf{Acc.} & \textbf{F\textsubscript{1}} 
& \textbf{Acc.} & \textbf{F\textsubscript{1}} 
& \textbf{Acc.} & \textbf{F\textsubscript{1}} 
& \textbf{Acc.} & \textbf{F\textsubscript{1}} 
& \textbf{Acc.} & \textbf{F\textsubscript{1}} 
& \textbf{Acc.} & \textbf{F\textsubscript{1}} \\
\midrule
\multicolumn{13}{l}{\textit{Baselines}} \\
\midrule
\textbf{\textsc{zero-shot}}     & 61.7 & 55.3 & 62.7 & 57.0 & 62.7 & 57.3  & \textbf{57.8} & \textbf{51.0} & 55.8 & 47.7 & 60.5 & 51.3 \\
\textbf{\textsc{def}}    & 62.1 & 58.7 & 65.0 & 58.7 & 62.2 & 56.5 & 57.5 & 50.6 & 59.1 & 51.5 & \textbf{63.5} & \textbf{55.2} \\
\textbf{\textsc{logical forms}}    & \textbf{63.2} & \textbf{57.4} & \textbf{65.4} & \textbf{59.4} & \textbf{63.1} & \textbf{55.4} & 57.8 & 49.4 & \textbf{60.2} & \textbf{51.3} & 62.8 & 53.9 \\
\midrule
\multicolumn{13}{l}{\textit{LLM-derived Patterns and Definitions}} \\
\midrule
\textbf{\textsc{new def}}     & 66.8 & 67.3 & 66.8 & 59.9 & 66.8 & 60.0 & 57.5 & 52.5 & 58.8 & 53.3 & 64.8 & 57.7 \\
\textbf{\textsc{patterns}}   & \textbf{72.2} & \textbf{66.4} & 73.2 & 64.9 & 70.5 & 66.2 & 63.5 & 55.7 & 64.5 & 53.3 & \textbf{68.5} & \textbf{61.9} \\
\textbf{\textsc{pattern matching}} & 70.1 & 65.9 & \underline{\textbf{73.5}} & \underline{\textbf{66.5}} & \textbf{71.5} & \textbf{66.5} & \textbf{65.2} & \textbf{57.9} & \textbf{66.2} & \textbf{59.6} & 67.2 & 59.9 \\
\midrule
\multicolumn{13}{l}{\textit{One-shot prompting}} \\
\midrule
\textbf{\textsc{one-shot}}    & 63.6 & 59.6 & 64.1 & 58.7 & 58.5 & 55.9 & 56.2 & 48.1 & 56.1 & 46.2 & 60.0 & 49.7\\
\textbf{\textsc{one-shot + exp}} & 65.2 & 59.5 & 63.5 & 59.0 & 45.7 & 48.6 & 56.8 & 50.0 & 56.3 & 47.9 & 59.2 & 49.7 \\
\textbf{\textsc{dynamic one-shot}} \\
\textit{all-MiniLM-L6-v2} & 70.2 & 67.6 & 71.3 & 66.4 & 70.4 & 66.2 & 65.8 & 61.7 & 65.5 & 59.7 & 68.5 & 63.3 \\
\textit{roberta-base} & 69.5 & 64.3 & 69.5 & 64.6 & 72.5 & 67.8 & 65.5 & 61.3 & 64.8 & 58.9 & 66.5 & 60.6 \\
\textbf{\textsc{syntax-based dynamic one-shot}} \\
\textit{all-MiniLM-L6-v2} & 68.2 & 63.6 & 71.2 & 66.1 & 68.5 & 64.2 & 63.2 & 58.3 & 62.8 & 55.7 & 64.5 & 57.3 \\
\textit{syntax-augmented roberta-large} & 65.5 & 65.5 & 71.2 & 66.3 & 68.5 & 64.2 & 64.5 & 58.6 & 64.2 & 56.5 & 63.5 & 56.0 \\
\textbf{\textsc{dynamic + exp}} & 71.2 & 68.9 & 69.5 & 65.0 & 72.7 & 67.9 & \underline{\textbf{67.8}} & \underline{\textbf{61.0}} & \underline{\textbf{67.5}} & \underline{\textbf{62.2}} & 68.2 & 63.2 \\
\textbf{\textbf{\textsc{dynamic + exp + patterns}}} & \underline{\textbf{74.2}} & \underline{\textbf{68.9}} & \textbf{73.1} & \textbf{67.2} & \underline{\textbf{73.2}} & \underline{\textbf{67.9}} & 66.8 & 62.3 & 67.2 & 55.1 & \underline{\textbf{70.5}} & \underline{\textbf{65.9}} \\
\midrule
\multicolumn{13}{l}{\textit{Multi-step classification}} \\
\midrule
\textbf{\textsc{multistep}} & 65.4 & 64.9 & 70.9 & 62.7 & 62.2 & 57.1 & 65.8 & 55.8 & 62.5 & 55.1 & 66.8 & 60.2 \\
\bottomrule
\end{tabular}
}
\caption{Fallacy classification performance on \textsc{Logic}. \textbf{Bold}: best approach in section per model by accuracy, \underline{\textbf{Bold}}: best approach overall per model by accuracy. F\textsubscript{1} score denotes Macro F\textsubscript{1} score, which accounts for the class imbalance in the dataset.}
\label{tab:logic-results-wide}
\end{table*}

Table~\ref{tab:logic-results-wide} summarizes all experimental configurations and results on \textsc{Logic}. It reveals a consistent improvement when the model leverages information about the underlying logic extracted through the LLMs, 
especially with reasoning models and \texttt{gpt-4o}.
When using reasoning models, the model-generated definitions yield a 4.65\% accuracy improvement over our manually corrected definitions. In the same way, including our generated patterns causes a 8.2\% increase with respect to the logical forms extracted by the website logicallyfallacious.com and used in \citet{jin2022logicalfallacydetection}. McNemar's test proved statistical significance for all models using \textsc{\textbf{patterns}} against \textsc{\textbf{zero-shot}} and for all except \texttt{llama} and \texttt{deepseek} against \textsc{\textbf{logical forms}} method. When it comes to non-reasoning models, the new definitions do not really affect the performance, whereas using our patterns improves the accuracy by 5.8\% on average. \eleni{For comparison, we test our method against \citet{robbani-etal-2024-flee}'s templates: our patterns outperform said schemes by an average 10.7\% across all models.}

A notable result is the performance increase achieved through dynamic one-shot prompting. In particular, \textbf{\textsc{dynamic one-shot}} approach (using \texttt{{}all-MiniLM-L6-v2}) yields an average $8.87\%$ increase in accuracy compared to \textbf{\textsc{one-shot}}, despite relying on semantic similarity for example selection. On the other hand, the syntax-oriented example retrieval strategy (\textbf{\textsc{syntax-based dynamic one-shot}}) does not outperform the semantic selection. This may be partially 
due to inaccuracies in the sentence masking process, which can negatively impact the retrieval of similar examples and the classification, consequently.
The \textbf{\textsc{multistep}} approach shows weaker performance than \textbf{\textsc{pattern matching}}, especially for \texttt{deepseek-r1}~\cite{liu2024deepseekv3}, implying that generating logical forms without explicit guidance constitutes the main challenge for the model in the request. 

\begin{table}[!b]
\renewcommand{\arraystretch}{0.9}
\centering
\setlength{\tabcolsep}{4pt} 
\scalebox{0.8}{
\begin{tabular}{lcc}
\toprule
\textbf{Method} & \textbf{Acc} & \textbf{F\textsubscript{1}} \\
\midrule
\texttt{\citet{jeong2025largelanguagemodelsbetter}} & 49.0 & 37.0 \\
\texttt{\citet{pan2024llmsgoodzeroshotfallacy}} & - & 50.5 \\
\textsc{\textbf{patterns}} (\texttt{gpt-4o}) & 73.5 & 66.5 \\
\textsc{\textbf{dynamic+exp+pat.}} (\texttt{o4-mini}) & 74.2 & 68.9 \\
\bottomrule
\end{tabular}%
}
\caption{Comparison of our best results against the unsupervised baselines provided by \citet{jeong2025largelanguagemodelsbetter} and \citet{pan2024llmsgoodzeroshotfallacy} (described in Appendix~\ref{sec:baseline}) for \textsc{Logic}. 
}
\label{tab:reduced-results}
\end{table}
In summary, 
including context-aware logical patterns proves consistently beneficial for fallacy classification: \textbf{\textsc{patterns}} with \texttt{gpt-4o} reaches 73.5\% accuracy, outperforming prior unsupervised methods 
(Table~\ref{tab:reduced-results}), while \textbf{\textsc{dynamic+exp+patterns}} with \texttt{o4-mini} achieves 74.2\% when augmented with examples and patterns.


\subsection{Error analysis}

\paragraph{Pattern matching} Requesting the model to identify the closest pattern for each argument provides insight into the association process between sentences and patterns. For our analysis, we have split our fallacies into two groups in Table~\ref{tab:fallacies-group}: \mbox{\textit{(i)} group 1}, consisting of fallacies whose patterns include logical forms while still including additional contextual cues; \textit{(ii)} group 2, consisting of fallacies that lack highly structured patterns and rely more on contextual and semantic features of the sentence.

\begin{table}[!b]
\centering
\small
\renewcommand{\arraystretch}{1}
\begin{tabular}{@{}p{3cm}p{3.8cm}@{}}
\toprule
\textbf{Group 1} & \textbf{Group 2} \\
\midrule
\begin{minipage}[t]{\linewidth}
\begin{itemize}[left=1em, nosep]
  \item Ad Hominem
  \item Ad Populum
  \item Circular Reasoning
  \item Irrelevant Authority
  \item False Cause
  \item Hasty Generalization
  \item Deductive Fallacy
  \item Black-and-White Fallacy
\end{itemize}
\end{minipage}
&
\begin{minipage}[t]{\linewidth}
\begin{itemize}[left=1em, nosep]
  \item Red Herring
  \item Equivocation
  \item Emotional Language
  \item Extension Fallacy
  \item Intentional Fallacy
\end{itemize}
\end{minipage} \\
\bottomrule
\end{tabular}
\caption{Grouped fallacy classes based on pattern features for analytical purposes.}
\label{tab:fallacies-group}
\end{table}

\begin{figure}[!h]
\centering
\begin{tikzpicture}
\begin{axis}[
    ybar,
    bar width=5pt,
    width=0.4\textwidth,
    height=4.5cm,
    enlarge x limits=0.2,
    ylabel={\% F\textsubscript{1} score},
    symbolic x coords={o4-mini,gpt-4o, deepseek-r1,gpt-4.1-mini,llama-3.3-70B,gemma-3-27B-it},
    xtick=data,
    xticklabel style={
    rotate=10,
    anchor=east,
    font=\scriptsize
    },
    ymin=0,
    ymax=100,
    legend style={
    font=\scriptsize,
    at={(1.05,0.5)},
    anchor=west,
    draw=none,
    cells={anchor=west},
    legend cell align=left
    },
    tick label style={font=\small},
    label style={font=\small},
    nodes near coords,
    every node near coord/.append style={font=\tiny},
    xticklabel style={rotate=13, anchor=north},
]

\addplot+[fill=blue!50] coordinates {(o4-mini,71) (gpt-4o, 74) (deepseek-r1, 73)(gpt-4.1-mini,65) (llama-3.3-70B,67) (gemma-3-27B-it,71)};
\addplot+[fill=red!60]  coordinates {(o4-mini,56) (gpt-4o, 54) (deepseek-r1, 56)(gpt-4.1-mini,39) (llama-3.3-70B,43) (gemma-3-27B-it,47)};

\legend{Group 1, Group 2}
\end{axis}
\end{tikzpicture}
\caption{Group-wise F\textsubscript{1} score for each model, relative to the \textbf{\textsc{pattern matching}} prompt setting.}
\label{fig:groupwisef1}
\end{figure}
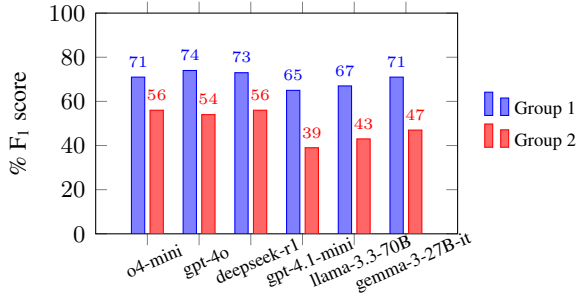

Figure~\ref{fig:groupwisef1} shows consistently superior accuracy 
for Group 1, whose classes maintain relatively high performance across all experimental settings. The class \textit{Circular Reasoning} emerges as the most accurately predicted class across all models. For what concerns Group 2, the overall accuracy 
is, on average, 22\% lower with respect to Group 1. 
The classes \textit{Emotional Language}, \textit{Red Herring} and \textit{Extension Fallacy} achieve moderate prediction accuracy, whereas only \textit{Evading the Burden of Proof}'s patterns within the \textit{Intentional Fallacy} category are correctly classified, and \textit{Equivocation} remains entirely undetected by \texttt{gpt-4.1-mini}~\cite{openai2023gpt4}. In summary, the models achieve better performance on logical fallacies that exhibit clearer structural characteristics but face difficulties with fallacies requiring more nuanced semantic understanding and contextual analysis.

\begin{table*}[h]
\renewcommand{\arraystretch}{1.05}
\centering
\fontsize{10.5pt}{12.5pt}\selectfont
\begin{tabular}{@{}m{0.10\textwidth} p{0.86\textwidth}@{}}
\toprule[1.1pt]
\multicolumn{2}{p{0.96\textwidth}@{}}{\cellcolor{gray!8}
I have no intention of stopping the use of somatostatin on patients suffering from acute pancreatitis. I consider it to be a very reasonable choice. After all, it has been standard practice in our department for many years and we’ve been quite satisfied with the results we’ve had. \textit{\textbf{Irrelevant Authority}}
} \\
\midrule[0.4pt]
\textbf{Top 1} &
Because many people [do/believe/support] X, X must be true/good/right/best/valid. (\textit{Ad Populum}) \\
\textbf{Top 2} &
Using [personal trait, experience, past action] as implicit proof of authority on a distinct or unrelated subject. (\textit{Irrelevant Authority}) \\
\bottomrule[1.1pt]
\end{tabular}
\caption{Sentence accurately classified by \texttt{o4-mini} with the $2^{nd}$ ranked pattern. Although the model fails initially, the matching can still be deemed reasonable.}
\label{tab:example-correction-top2}
\end{table*}

Furthermore, matching patterns allows us to see that some instances can be deemed as fitting from a structural point of view
, thus partially explaining the inherent difficulty of the classification task. While providing guidance through syntactic and logical structure proves beneficial for fallacy detection, this approach does not eliminate all sources of ambiguity, as some sentences may conform to multiple structural patterns. The critical point lies in context-aware pattern application. 
Models must not only identify logical forms but also evaluate their contextual validity in each sentence.

To quantify the degree of ambiguity inherent in pattern matching, we instructed the best-performing model \texttt{o4-mini} to return the five most similar patterns for each argument. This multi-candidate approach enables us to analyze whether lower-ranked patterns might also represent valid interpretations of the same argument. By examining the distribution of pattern similarities and evaluating classification accuracy when considering alternative matches, we can better understand the boundaries of pattern-based classification and identify instances where structural ambiguity genuinely complicates fallacy detection.

Table~\ref{tab:top-5} shows that, when the model is prompted to return multiple matching patterns rather than a single best match, its confidence in the initial prediction decreases, resulting in a 3.4\% drop in accuracy 
(see Table~\ref{tab:logic-results-wide}). 

\begin{table}[H]
\centering
\setlength{\tabcolsep}{3pt} 
\scalebox{0.85}{
\begin{tabular}{@{}ccccc@{}}
\toprule
\textbf{Acc@1} & \textbf{Acc@2} & \textbf{Acc@3} & \textbf{Acc@4} & \textbf{Acc@5} \\ \midrule
66.7         & 75.1         & 81.8         & 86.5         & 88.5         \\ \bottomrule
\end{tabular}
}
\caption{Performance analysis in \textbf{\textsc{pattern matching}} with expanded solution pool: classification results including top 5 predictions as correct.}
\label{tab:top-5}
\end{table}

However, this apparent degradation is misleading when viewed in isolation. By incorporating the second-ranked pattern choice into our evaluation, performance recovers to $75.1\%$, and continues to improve as we expand our candidate pool to include progressively lower-ranked options. Table~\ref{tab:example-correction-top2} illustrates a representative case where the model successfully identifies the correct pattern as its second choice, while its first-ranked selection remains structurally plausible. The model likely assigns one of the \textit{Ad Populum} patterns because it closely matches the argument's logic, while the \textit{Irrelevant Authority} pattern does not fit the sentence since it requires discussion of an unrelated topic, which is not present in the sentence.  These subtle distinctions likely make pattern matching more challenging than direct classification because it requires strict structural alignment as well as capturing broader content-related features.

\begin{table*}[!h]
\renewcommand{\arraystretch}{1.05}
\centering
\fontsize{10.5pt}{12.5pt}\selectfont
\begin{tabular}{@{}m{0.16\textwidth} m{0.80\textwidth}@{}}
\toprule[1.1pt]
\multicolumn{2}{p{\dimexpr 0.96\textwidth\relax}@{}}{\cellcolor{gray!8}
Teaching assistant: the homework assignment was much harder than we thought, so I think we should give a few extra points to students who completed it. Professor: that's a terrible idea. If we give everyone a perfect score for no reason, students won't bother working hard in the future. \textit{\textbf{Extension Fallacy}}
} \\
\midrule[0.4pt]
\centering\texttt{o4-mini} &
\textbullet~\textbf{Generated Form:} A proposes: ``Because B, we should do $X_1$.'' Opponent replies: ``If we do $X_2$ (an exaggerated version of $X_1$) for no reason, then C will happen; therefore, we must not do $X_1$.'' \newline
\textbullet~\textbf{Matched Pattern:} X proposes Y; response exaggerates Y to an extreme version Y. \newline
\textbullet~\textbf{Fallacy:} Extension Fallacy \\
\midrule[0.4pt]
\centering\texttt{gpt-4.1-mini} &
\textbullet~\textbf{Generated Form:} A says B was harder than expected, so A suggests giving extra points to those who completed it. B replies: if we give everyone a perfect score for no reason, then C won't work hard; therefore, we should not give extra points. \newline
\textbullet~\textbf{Matched Pattern:} Arguing that taking a minor action will cause a chain of escalating events leading to a disastrous outcome, even though no strong evidence supports the inevitability of that chain. \newline
\textbullet~\textbf{Fallacy:} Hasty Generalization \\
\midrule[0.4pt]
\centering\texttt{llama-3.3-70B} &
\textbullet~\textbf{Generated Form:} If we do A (give everyone a perfect score for no reason), then B (students won't bother working hard in the future). \newline
\textbullet~\textbf{Matched Pattern:} Presenting a coincidental or correlated event as proof of causation, often ignoring other plausible causal factors or explanations. \newline
\textbullet~\textbf{Fallacy:} False Cause \\
\midrule[0.4pt]
\centering\texttt{gemma-3-27b-it} &
\textbullet~\textbf{Generated Form:} If we do A (give extra points), then B (students won't work hard). Therefore, we should not do A. \newline
\textbullet~\textbf{Matched Pattern:} If P then Q; Q is true; therefore, P is true. \newline
\textbullet~\textbf{Fallacy:} Deductive Fallacy \\
\bottomrule[1.1pt]
\end{tabular}
\caption{Comparison of outputs from four models evaluated in the \textbf{\textsc{multistep}} configuration on \textsc{Logic}.}
\label{tab:multi-step-output}
\end{table*}

\paragraph{Multistep classification.} The \textbf{\textsc{multistep}} approach fails to produce significant results. We conduct this experiment in a single passage to force the model to reason using both semantic and syntactic information. However, classification performance depends critically on the quality of the extracted logical forms, which proves inconsistent and model-dependent. For instance, \texttt{o4-mini} embeds classification-relevant contextual information directly into its generated logical forms (Table~\ref{tab:multi-step-output}). 
Furthermore, models demonstrate substantially weaker performance on Group 2 sentences compared to Group 1, showing an average decrease of 21.5\% in F\textsubscript{1} score. Additionally, models frequently bypass the pattern matching phase entirely, arbitrarily assigning 
patterns despite clear misalignment with the extracted logical form. For example, given the argument \textit{People nowadays only vote with their emotions instead of their brains} (an instance of \textit{Hasty Generalization}), the model \texttt{o4-mini} first extracts the logical form \textit{All A only do B instead of C}. The model then matches this form to the pattern \textit{Generalizing from a small sample or single event to an entire group or population}, which correctly belongs to \textit{Hasty Generalization}. While this produces an accurate classification, the assigned pattern does not precisely correspond to the extracted logical form. In summary, while humans naturally decompose pattern matching into multiple cognitive steps, this multi-stage process proves to be challenging for current LLMs. Models struggle to bridge the gap between abstract logical patterns and their content-dependent manifestations, often failing to identify the implicit premises and unstated logical connections that underlie the reasoning chain.

\begin{table}[!ht]
\small
\renewcommand{\arraystretch}{1}
\centering
\resizebox{\columnwidth}{!}{%
\begin{tabular}{@{}lllll@{}}
\toprule
                                  & \multicolumn{2}{l}{\textbf{\textsc{Reddit}}} & \multicolumn{2}{l}{\textbf{\textsc{ElecDebate}}} \\ \midrule
                                  & \textbf{Acc.}        & \textbf{M-F1\textsubscript{1}}        & \textbf{Acc.}        & \textbf{M- F1\textsubscript{1}}   \\ \midrule
\scriptsize{\textbf{\textsc{zero-shot}}}      & 82.8           & 82.8               & 67.3          & 50.8   \\
\scriptsize{\textbf{\textsc{def}}}             & 82.6           & 82.5               & 65.9          & 54.7   \\
\scriptsize{\textbf{\textsc{logical forms}}}   & 84.7           & 84.3               & 70.7          & 59.5   \\
\scriptsize{\textbf{\textsc{patterns}}}        & \textbf{84.7}  & \textbf{84.5}      & 65.5          & 56.3   \\
\scriptsize{\textbf{\textsc{pattern matching}}}& 80.9           & 80.8               & 65.5          & 56.7   \\
\scriptsize{\textbf{\textsc{dynamic one-shot}}}& 81.9           & 81.6               & \textbf{81.7} & \textbf{70.4}   \\
\scriptsize{\textbf{\textsc{dynamic + exp}}}   & 83.8           & 83.6               & 79.1          & 71.3   \\
\scriptsize{\textbf{\textsc{dynamic + exp + patterns}}} & 79.0  & 78.8               & 78.8          & 72.3   \\
\midrule
\scriptsize{\textbf{\textsc{same-dataset patterns}}} & 83.8        & 83.4                  & 74.1          & 64.9   \\
\scriptsize{\textbf{\textsc{same-dataset patterns matching}}} & 84.7        & 84.3                  & 74.3          & 63.7  \\

\bottomrule
\end{tabular}%
}
\caption{Fallacy classification performance using \texttt{o4-mini} on \textsc{Reddit} and \textsc{ElecDebate}. \textbf{\textsc{patterns}} method involves using patterns generated on \textsc{Logic} while \textbf{\textsc{same-dataset patterns}} approach includes patterns generated on the datasets \textsc{Reddit} and \textsc{ElecDebate} themselves. Acc.: Accuracy; M-F1: Macro-F1.
}
\label{tab:other-datasets}
\end{table}

\section{Experiments on Further Datasets} 
\label{sec:pattern-evaluation}
In order to further assess the quality of \textsc{Logic}-derived patterns, we conducted a subset of the experiments on  \textsc{Reddit} and \textsc{ElecDebate} 
using the best performing model, \texttt{o4-mini}.
We tested patterns extracted from \textsc{Logic}, restricted to the two datasets' classes (first eight rows in Table~\ref{tab:other-datasets}) and patterns extracted from the datasets themselves (latest two rows in Table~\ref{tab:other-datasets}). \textsc{Reddit} patterns show a prevalence of linguistic markers over logical forms while \textsc{ElecDebate} ones emphasize stronger logical formalization, incorporating symbolic formalism.

Consistent with previous findings, logical pattern incorporation outperforms competing approaches on \textsc{Reddit}. Moreover, \textsc{Logic}-based and \textsc{Reddit}-based patterns yield comparable results.
While taxonomy alignment prevents direct comparison, results from supervised and unsupervised methods \citep{sahai-etal-2021-breaking, lei2024boostinglogicalfallacyreasoning,pan2024llmsgoodzeroshotfallacy, yeh2024cocolofadatasetnewscomments} are consistent with our findings (see Appendix~\ref{sec:baseline}). 
Only a comparison with \citet{lei2024boostinglogicalfallacyreasoning} (Macro F\textsubscript{1}=81.3\%) is possible: \textsc{\textbf{patterns}} and \textsc{\textbf{same-dataset patterns}} outperform their results.

Regarding \textsc{ElecDebate}, Table~\ref{tab:other-datasets} shows that \textbf{\textsc{dynamic one-shot}} yields the best performance, possibly due to the predominant presence of the class \textit{Emotional Language} (62.5\% of test set) whose detection may particularly benefit from similar worded examples. 
Indeed, 
\textbf{\textsc{same-dataset patterns}} 
achieve competitive results 
with respect to \citet{goffredo-etal-2023-argument, pan2024llmsgoodzeroshotfallacy} (see Appendix~\ref{sec:baseline}).
These experiments showed a fair generalization of \textsc{Logic}-derived patterns on other datasets, with the additional advantage of not requiring labeled data to re-extract the patterns. 

In order to prove the broader applicability of our approach beyond \textsc{Logic}-specific patterns, we tested patterns generated from the other two datasets on \textsc{Logic}. 

Table~\ref{tab:other-datasets-on-logic} demonstrates solid results, validating findings on \textsc{Logic} and proving the robustness and transferability of our pattern-based methodology.
Notice that accuracy is not directly comparable with values in Table~\ref{tab:logic-results-wide} since \textsc{Reddit} and \textsc{ElecDebate} have a subset of the classes of \textsc{Logic}.





\begin{table}[H]
\small
\renewcommand{\arraystretch}{1}
\centering
\resizebox{\columnwidth}{!}{%
\begin{tabular}{@{}lll@{}}
\toprule
                                  & \textbf{\textsc{Logic\textsubscript{Reddit}}} & \textbf{\textsc{Logic\textsubscript{ElecDebate}}} \\ \midrule
                                  & \textbf{Acc.}        & \textbf{Acc.}      \\ \midrule
\scriptsize{\textbf{\textsc{same-dataset patterns}}} & 90.8        & 87.1             \\
\scriptsize{\textbf{\textsc{same-dataset patterns matching}}} & 88.3        & 87.5    \\
\scriptsize{\textbf{\textsc{Logic patterns}}} & 89.1        & 83.7             \\
\scriptsize{\textbf{\textsc{Logic patterns matching}}} & 90.0        & 87.1    \\
\bottomrule
\end{tabular}%
}
\caption{Fallacy classification performance using \texttt{o4-mini} on \textsc{Logic}. \textsc{Logic\textsubscript{X}} refers to \textsc{Logic} restricted to the classes from dataset \textsc{X}. \textbf{\textsc{same-dataset patterns}} approach includes patterns generated on non-\textsc{Logic} dataset \textsc{X} while \textbf{\textsc{Logic patterns}} involves using \textsc{Logic}-derived patterns restricted to the classes of dataset \textsc{X}.}
\label{tab:other-datasets-on-logic}
\end{table}

\section{Conclusions}
Fallacy detection is a challenging yet critical task to solve. 
\eleni{Since fallacies often manifest in nuanced and context-dependent forms, purely abstract representations are insufficient to characterize the full spectrum of ways a fallacy can appear in natural language, thus motivating the need to combine logical structure with context-level linguistic cues.} We present an experimental framework that inductively extracts context-aware structural patterns from fallacious arguments and their explanations, demonstrating that incorporating such patterns significantly enhances fallacy classification performance. Specifically, pattern-based classification achieves 73.5\% accuracy on \textsc{Logic}, significantly outperforming prior unsupervised approaches, and 74.2\% including one-shot examples. \eleni{Being data-driven, these patterns are not bound to a fixed set of fallacies and can flexibly capture the diverse nuances through which each fallacy type manifests.} Notably, reasoning models demonstrate consistently superior performance across all experimental configurations.
Moreover, experiments on additional datasets confirm that the extracted patterns generalize effectively across domains, establishing data-driven pattern extraction as an effective method to generate 
generalizable logical representations.  

\section{Limitations}
While this work demonstrates the efficacy of large language models in detecting logical fallacies by exploiting the underlying logical structure of sentences, it has several limitations. First, we intentionally generated patterns exclusively from the \textsc{Logic} dataset due to the quality and straightforward structure of its sentences. We are aware, however, that it does not fully cover the complex and multi-faceted spectrum of fallacies.
Furthermore, our work is based on a small sample of LLMs. Nevertheless, we selected a diverse and representative subset, including models from different providers, with varying sizes and reasoning capabilities.

\section{Ethics Statement}
Logical fallacies can reinforce societal bias and facilitate the spread of misinformation, leading to harmful consequences for society. This work focuses on leveraging LLMs for detecting logical fallacies in argumentation and should not be employed to manipulate discourse by exploiting identified reasoning patterns. Furthermore, this approach risks amplifying existing LLM biases, potentially causing unfair detection. We acknowledge these limitations and encourage future bias mitigation research. 
We are aware of the environmental impact of large-scale LLMs usage. However, this study exclusively employs inference-only methods, significantly reducing computational requirements compared to training approaches.
All datasets are used in accordance with their license and they have been checked for personally identifying and offensive content. 

\section*{Acknowledgements}

This publication is part of the project PNRR-NGEU, which has received funding from the MUR - DM 629/2024. 
We would like to thank the Qatar National Research Fund, part of Qatar Research Development and Innovation Council (QRDI), for also funding this work  by grant NPRP14C0916-210015.

\bibliography{anthology,custom}

@article{openai2023gpt4,
  title={{GPT-4} Technical Report},
  author={{OpenAI}},
  journal={arXiv preprint arXiv:2303.08774},
  year={2023}
}

@article{dubey2024llama,
  title={The {Llama} 3 herd of models},
  author={Dubey, Abhimanyu and Jauhri, Abhinav and Pandey, Abhinav and Kadian, Abhishek and Al-Dahle, Ahmad and Letman, Aiesha and Mathur, Akhil and Schelten, Alan and Yang, Amy and Fan, Angela and others},
  journal={arXiv preprint arXiv:2407.21783},
  year={2024}
}

@article{liu2024deepseekv3,
  title         = {{DeepSeek-V3} Technical Report},
  author        = {{DeepSeek-AI} and Liu, Aixin and
                   Feng, Bei and
                   Xue, Bing and
                   Wang, Bingxuan and
                   Wu, Bochao and
                   Lu, Chengda and
                   Zhao, Chenggang and
                   Deng, Chengqi and
                   Zhang, Chenyu and
                   Ruan, Chong and
                   Dai, Damai and
                   Guo, Daya and
                   Yang, Dejian and
                   Chen, Deli and
                   Ji, Dongjie and
                   Li, Erhang and
                   Lin, Fangyun and
                   Dai, Fucong and
                   Luo, Fuli and
                   Hao, Guangbo and
                   Chen, Guanting and
                   Li, Guowei and
                   Zhang, H. and
                   Bao, Han and
                   Xu, Hanwei and
                   Wang, Haocheng and
                   Zhang, Haowei and
                   Ding, Honghui and
                   Xin, Huajian and
                   Gao, Huazuo and
                   Li, Hui and
                   Qu, Hui and
                   Cai, J.L. and
                   Liang, Jian and
                   Guo, Jianzhong and
                   Ni, Jiaqi and
                   Li, Jiashi and
                   Wang, Jiawei and
                   Chen, Jin and
                   Chen, Jingchang and
                   Yuan, Jingyang and
                   Qiu, Junjie and
                   Li, Junlong and
                   Song, Junxiao and
                   Dong, Kai and
                   Hu, Kai and
                   Gao, Kaige and
                   Guan, Kang and
                   Huang, Kexin and
                   Yu, Kuai and
                   Wang, Lean and
                   Zhang, Lecong and
                   Xu, Lei and
                   Xia, Leyi and
                   Zhao, Liang and
                   Wang, Litong and
                   Zhang, Liyue and
                   Li, Meng and
                   Wang, Miaojun and
                   Zhang, Mingchuan and
                   Zhang, Minghua and
                   Tang, Minghui and
                   Li, Mingming and
                   Tian, Ning and
                   Huang, Panpan and
                   Wang, Peiyi and
                   Zhang, Peng and
                   Wang, Qiancheng and
                   Zhu, Qihao and
                   Chen, Qinyu and
                   Du, Qiushi and
                   Chen, R.J. and
                   Jin, R.L. and
                   Ge, Ruiqi and
                   Zhang, Ruisong and
                   Pan, Ruizhe and
                   Wang, Runji and
                   Xu, Runxin and
                   Zhang, Ruoyu and
                   Chen, Ruyi and
                   Li, S.S. and
                   Lu, Shanghao and
                   Zhou, Shangyan and
                   Chen, Shanhuang and
                   Wu, Shaoqing and
                   Ye, Shengfeng and
                   Ma, Shirong and
                   Wang, Shiyu and
                   Zhou, Shuang and
                   Yu, Shuiping and
                   Zhou, Shunfeng and
                   Pan, Shuting and
                   Wang, T. and
                   Yun, Tao and
                   Pei, Tian and
                   Sun, Tianyu and
                   Xiao, W.L. and
                   Zeng, Wangding and
                   Zhao, Wanjia and
                   An, Wei and
                   Liu, Wen and
                   Liang, Wenfeng and
                   Gao, Wenjun and
                   Yu, Wenqin and
                   Zhang, Wentao and
                   Li, X.Q. and
                   Jin, Xiangyue and
                   Wang, Xianzu and
                   Bi, Xiao and
                   Liu, Xiaodong and
                   Wang, Xiaohan and
                   Shen, Xiaojin and
                   Chen, Xiaokang and
                   Zhang, Xiaokang and
                   Chen, Xiaosha and
                   Nie, Xiaotao and
                   Sun, Xiaowen and
                   Wang, Xiaoxiang and
                   Cheng, Xin and
                   Liu, Xin and
                   Xie, Xin and
                   Liu, Xingchao and
                   Yu, Xingkai and
                   Song, Xinnan and
                   Shan, Xinxia and
                   Zhou, Xinyi and
                   Yang, Xinyu and
                   Li, Xinyuan and
                   Su, Xuecheng and
                   Lin, Xuheng and
                   Li, Y.K. and
                   Wang, Y.Q. and
                   Wei, Y.X. and
                   Zhu, Y.X. and
                   Zhang, Yang and
                   Xu, Yanhong and
                   Huang, Yanping and
                   Li, Yao and
                   Zhao, Yao and
                   Sun, Yaofeng and
                   Li, Yaohui and
                   Wang, Yaohui and
                   Yu, Yi and
                   Zheng, Yi and
                   Zhang, Yichao and
                   Shi, Yifan and
                   Xiong, Yiliang and
                   He, Ying and
                   Tang, Ying and
                   Piao, Yishi and
                   Wang, Yisong and
                   Tan, Yixuan and
                   Ma, Yiyang and
                   Liu, Yiyuan and
                   Guo, Yongqiang and
                   Wu, Yu and
                   Ou, Yuan and
                   Zhu, Yuchen and
                   Wang, Yuduan and
                   Gong, Yue and
                   Zou, Yuheng and
                   He, Yujia and
                   Zha, Yukun and
                   Xiong, Yunfan and
                   Ma, Yunxian and
                   Yan, Yuting and
                   Luo, Yuxiang and
                   You, Yuxiang and
                   Liu, Yuxuan and
                   Zhou, Yuyang and
                   Wu, Z.F. and
                   Ren, Z.Z. and
                   Ren, Zehui and
                   Sha, Zhangli and
                   Fu, Zhe and
                   Xu, Zhean and
                   Huang, Zhen and
                   Zhang, Zhen and
                   Xie, Zhenda and
                   Zhang, Zhengyan and
                   Hao, Zhewen and
                   Gou, Zhibin and
                   Ma, Zhicheng and
                   Yan, Zhigang and
                   Shao, Zhihong and
                   Xu, Zhipeng and
                   Wu, Zhiyu and
                   Zhang, Zhongyu and
                   Li, Zhuoshu and
                   Gu, Zihui and
                   Zhu, Zijia and
                   Liu, Zijun and
                   Li, Zilin and
                   Xie, Ziwei and
                   Song, Ziyang and
                   Gao, Ziyi and
                   Pan, Zizheng},
  journal       = {arXiv preprint arXiv:2412.19437},
  year          = {2024},
  eprint        = {2412.19437},
  archivePrefix = {arXiv}
}

@book{Copi1953-COPITL-9,
	address = {New York, NY, USA},
	author = {Irving Marmer Copi and Carl Cohen and Kenneth McMahon},
        publisher = {Macmillan},
	editor = {Carl Cohen and K. D. McMahon},
	title = {Introduction to Logic},
	year = {1953}
}

@book{bacon1999logic,
  title={Logic from A to Z: The Routledge Encyclopedia of Philosophy Glossary of Logical and Mathematical Terms},
  author={Bacon, John B. and Detlefsen, Michael and McCarty, David Charles},
  year={1999},
  publisher={Routledge},
  doi={10.4324/9780203754696}
}

@book{hamblin1970fallacies,
  title={Fallacies},
  author={Hamblin, C.L.},
  isbn={9780416145700},
  lccn={lc71466947},
  series={University paperbacks},
  url={https://books.google.it/books?id=bYYIAQAAIAAJ},
  year={1970},
  publisher={Methuen}
}

@book{Johnson1977-JOHLS,
	address = {Toronto, Canada},
	author = {Ralph Henry Johnson and J. Anthony Blair},
	editor = {},
	title = {Logical Self-Defense},
	year = {1977}
}

@inproceedings{jin2022logicalfallacydetection,
    title = "Logical Fallacy Detection",
    author = {Jin, Zhijing  and
      Lalwani, Abhinav  and
      Vaidhya, Tejas  and
      Shen, Xiaoyu  and
      Ding, Yiwen  and
      Lyu, Zhiheng  and
      Sachan, Mrinmaya  and
      Mihalcea, Rada  and
      Sch{\"o}lkopf, Bernhard},
    editor = "Goldberg, Yoav  and
      Kozareva, Zornitsa  and
      Zhang, Yue",
    booktitle = "Findings of the Association for Computational Linguistics: EMNLP 2022",
    month = dec,
    year = "2022",
    address = "Abu Dhabi, United Arab Emirates",
    publisher = "Association for Computational Linguistics",
    url = "https://aclanthology.org/2022.findings-emnlp.532/",
    doi = "10.18653/v1/2022.findings-emnlp.532",
    pages = "7180--7198"
}

@inproceedings{jeong2025largelanguagemodelsbetter,
    title = "Large Language Models Are Better Logical Fallacy Reasoners with Counterargument, Explanation, and Goal-Aware Prompt Formulation",
    author = "Jeong, Jiwon  and
      Jang, Hyeju  and
      Park, Hogun",
    editor = "Chiruzzo, Luis  and
      Ritter, Alan  and
      Wang, Lu",
    booktitle = "Findings of the Association for Computational Linguistics: NAACL 2025",
    month = apr,
    year = "2025",
    address = "Albuquerque, New Mexico",
    publisher = "Association for Computational Linguistics",
    url = "https://aclanthology.org/2025.findings-naacl.384/",
    doi = "10.18653/v1/2025.findings-naacl.384",
    pages = "6933--6952",
    ISBN = "979-8-89176-195-7"
}

@inproceedings{hong2024closerlookselfverificationabilities,
    title = "A Closer Look at the Self-Verification Abilities of Large Language Models in Logical Reasoning",
    author = "Hong, Ruixin  and
      Zhang, Hongming  and
      Pang, Xinyu  and
      Yu, Dong  and
      Zhang, Changshui",
    editor = "Duh, Kevin  and
      Gomez, Helena  and
      Bethard, Steven",
    booktitle = "Proceedings of the 2024 Conference of the North American Chapter of the Association for Computational Linguistics: Human Language Technologies (Volume 1: Long Papers)",
    month = jun,
    year = "2024",
    address = "Mexico City, Mexico",
    publisher = "Association for Computational Linguistics",
    url = "https://aclanthology.org/2024.naacl-long.52/",
    doi = "10.18653/v1/2024.naacl-long.52",
    pages = "900--925"
}

@inproceedings{alhindi-etal-2024-large,
    title = "Large Language Models are Few-Shot Training Example Generators: A Case Study in Fallacy Recognition",
    author = "Alhindi, Tariq  and
      Muresan, Smaranda  and
      Nakov, Preslav",
    editor = "Ku, Lun-Wei  and
      Martins, Andre  and
      Srikumar, Vivek",
    booktitle = "Findings of the Association for Computational Linguistics: ACL 2024",
    month = aug,
    year = "2024",
    address = "Bangkok, Thailand",
    publisher = "Association for Computational Linguistics",
    url = "https://aclanthology.org/2024.findings-acl.732/",
    doi = "10.18653/v1/2024.findings-acl.732",
    pages = "12323--12334"
}

@inproceedings{pan2024llmsgoodzeroshotfallacy,
    title = "Are {LLM}s Good Zero-Shot Fallacy Classifiers?",
    author = "Pan, Fengjun  and
      Wu, Xiaobao  and
      Li, Zongrui  and
      Luu, Anh Tuan",
    editor = "Al-Onaizan, Yaser  and
      Bansal, Mohit  and
      Chen, Yun-Nung",
    booktitle = "Proceedings of the 2024 Conference on Empirical Methods in Natural Language Processing",
    month = nov,
    year = "2024",
    address = "Miami, Florida, USA",
    publisher = "Association for Computational Linguistics",
    url = "https://aclanthology.org/2024.emnlp-main.794/",
    doi = "10.18653/v1/2024.emnlp-main.794",
    pages = "14338--14364"
}

@inproceedings{sourati2023casebasedreasoninglanguagemodels,
  title     = {Case-Based Reasoning with Language Models for Classification of Logical Fallacies},
  author    = {Sourati, Zhivar and Ilievski, Filip and Sandlin, Hông-Ân and Mermoud, Alain},
  booktitle = {Proceedings of the Thirty-Second International Joint Conference on
               Artificial Intelligence, {IJCAI-23}},
  publisher = {International Joint Conferences on Artificial Intelligence Organization},
  editor    = {Edith Elkind},
  pages     = {5188--5196},
  year      = {2023},
  month     = {8},
  note      = {Main Track},
  doi       = {10.24963/ijcai.2023/576},
  url       = {https://doi.org/10.24963/ijcai.2023/576},
}

@inproceedings{reimers-2019-sentence-bert,
    title = "Sentence-BERT: Sentence Embeddings using Siamese BERT-Networks",
    author = "Reimers, Nils and Gurevych, Iryna",
    booktitle = "Proceedings of the 2019 Conference on Empirical Methods in Natural Language Processing",
    month = "11",
    year = "2019",
    publisher = "Association for Computational Linguistics",
    url = "http://arxiv.org/abs/1908.10084",
}

@inproceedings{lim2024evaluationllmidentifyinglogical,
author = {Lim, Gionnieve and Perrault, Simon T.},
title = {Evaluation of an LLM in Identifying Logical Fallacies: A Call for Rigor When Adopting LLMs in HCI Research},
year = {2024},
isbn = {9798400711145},
publisher = {Association for Computing Machinery},
address = {New York, NY, USA},
url = {https://doi.org/10.1145/3678884.3681867},
doi = {10.1145/3678884.3681867},
booktitle = {Companion Publication of the 2024 Conference on Computer-Supported Cooperative Work and Social Computing},
pages = {303–308},
numpages = {6},
location = {San Jose, Costa Rica},
series = {CSCW Companion '24}
}

@inproceedings{lei2024boostinglogicalfallacyreasoning,
    title = "Boosting Logical Fallacy Reasoning in {LLM}s via Logical Structure Tree",
    author = "Lei, Yuanyuan  and
      Huang, Ruihong",
    editor = "Al-Onaizan, Yaser  and
      Bansal, Mohit  and
      Chen, Yun-Nung",
    booktitle = "Proceedings of the 2024 Conference on Empirical Methods in Natural Language Processing",
    month = nov,
    year = "2024",
    address = "Miami, Florida, USA",
    publisher = "Association for Computational Linguistics",
    url = "https://aclanthology.org/2024.emnlp-main.730/",
    doi = "10.18653/v1/2024.emnlp-main.730",
    pages = "13157--13173"
}

@article{xu2025socratessmartypantstestinglogic, title={Socrates or Smartypants: Testing Logic Reasoning Capabilities of Large Language Models with Logic Programming-Based Test Oracles}, volume={40}, url={https://ojs.aaai.org/index.php/AAAI/article/view/39021}, DOI={10.1609/aaai.v40i23.39021}, abstractNote={Large Language Models (LLMs) have achieved significant progress in language understanding and reasoning. Evaluating and analyzing their logical reasoning abilities has therefore become essential. However, existing datasets and benchmarks are often limited to overly simplistic, unnatural, or contextually constrained examples. In response to the growing demand, we introduce SMARTYPAT-BENCH, a challenging, naturally expressed, and systematically labeled benchmark derived from real-world high-quality Reddit posts containing subtle logical fallacies. Unlike existing datasets and benchmarks, it provides more detailed annotations of logical fallacies and features more diverse data. To further scale up the study and address the limitations of manual data collection and labeling, such as fallacy-type imbalance and labor-intensive annotation, we introduce SMARTYPAT, an automated framework powered by logic programming-based oracles. SMARTYPAT utilizes Prolog rules to systematically generate logically fallacious statements, which are then refined into fluent natural language sentences by LLMs, ensuring precise fallacy rep- resentation. Extensive evaluation demonstrates that SMARTYPAT produces fallacies comparable in subtlety and quality to human-generated content and significantly outperforms baseline methods. Finally, experiments reveal insights into LLM capabilities, highlighting that while excessive reasoning steps hinder fallacy detection accuracy, structured reasoning enhances fallacy categorization performance.}, number={23}, journal={Proceedings of the AAAI Conference on Artificial Intelligence}, author={Xu, Zihao and Ding, Junchen and Lou, Yiling and Zhang, Kun and Gong, Dong and Li, Yuekang}, year={2026}, month={Mar.}, pages={19433–19440} }

@article{Storer_1949, 
        title={Carl G. Hempel and Paul Oppenheim. Studies in the logic of explanation. Philosophy of science, vol. 15 (1948), pp. 135–175.}, 
        volume={14}, 
        DOI={10.2307/2266531}, 
        number={2}, 
        journal={Journal of Symbolic Logic}, 
        author={Storer, Thomas}, 
        year={1949}, 
        pages={133–133}
}

@misc{o4,
  title={OpenAI o3 and o4-mini System Card},
  author={OpenAI},
  year={2025}}

@inproceedings{sachan2021syntaxtreeshelppretrained,
    title = "Do Syntax Trees Help Pre-trained Transformers Extract Information?",
    author = "Sachan, Devendra  and
      Zhang, Yuhao  and
      Qi, Peng  and
      Hamilton, William L.",
    editor = "Merlo, Paola  and
      Tiedemann, Jorg  and
      Tsarfaty, Reut",
    booktitle = "Proceedings of the 16th Conference of the European Chapter of the Association for Computational Linguistics: Main Volume",
    month = apr,
    year = "2021",
    address = "Online",
    publisher = "Association for Computational Linguistics",
    url = "https://aclanthology.org/2021.eacl-main.228/",
    doi = "10.18653/v1/2021.eacl-main.228",
    pages = "2647--2661"
}

@InProceedings{llm4lfd,
author="Teo, Nicole
and Huang, Donghao
and Cambria, Erik
and Wang, Zhaoxia",
editor="Yuan, Shuhan
and Malliaros, Fragkiskos
and Zheng, Xin",
title="Large Language Models for Logical Fallacy Detection",
booktitle="Trends and Applications in Knowledge Discovery and Data Mining",
year="2025",
publisher="Springer Nature Singapore",
address="Singapore",
pages="387--398",
isbn="978-981-96-8197-6"
}

@inproceedings{li2024reasonfallacyenhancinglarge,
    title = "Reason from Fallacy: Enhancing Large Language Models' Logical Reasoning through Logical Fallacy Understanding",
    author = "Li, Yanda  and
      Wang, Dixuan  and
      Liang, Jiaqing  and
      Jiang, Guochao  and
      He, Qianyu  and
      Xiao, Yanghua  and
      Yang, Deqing",
    editor = "Duh, Kevin  and
      Gomez, Helena  and
      Bethard, Steven",
    booktitle = "Findings of the Association for Computational Linguistics: NAACL 2024",
    month = jun,
    year = "2024",
    address = "Mexico City, Mexico",
    publisher = "Association for Computational Linguistics",
    url = "https://aclanthology.org/2024.findings-naacl.192/",
    doi = "10.18653/v1/2024.findings-naacl.192",
    pages = "3053--3066"
}

@inproceedings{NEURIPS2020_1457c0d6,
	author = {Brown, Tom and Mann, Benjamin and Ryder, Nick and Subbiah, Melanie and Kaplan, Jared D and Dhariwal, Prafulla and Neelakantan, Arvind and Shyam, Pranav and Sastry, Girish and Askell, Amanda and Agarwal, Sandhini and Herbert-Voss, Ariel and Krueger, Gretchen and Henighan, Tom and Child, Rewon and Ramesh, Aditya and Ziegler, Daniel and Wu, Jeffrey and Winter, Clemens and Hesse, Chris and Chen, Mark and Sigler, Eric and Litwin, Mateusz and Gray, Scott and Chess, Benjamin and Clark, Jack and Berner, Christopher and McCandlish, Sam and Radford, Alec and Sutskever, Ilya and Amodei, Dario},
	booktitle = {Advances in Neural Information Processing Systems},
	editor = {H. Larochelle and M. Ranzato and R. Hadsell and M.F. Balcan and H. Lin},
	pages = {1877--1901},
	publisher = {Curran Associates, Inc.},
	title = {Language Models are Few-Shot Learners},
	url = {https://proceedings.neurips.cc/paper_files/paper/2020/file/1457c0d6bfcb4967418bfb8ac142f64a-Paper.pdf},
	volume = {33},
	year = {2020}}

@article{wei2023chainofthoughtpromptingelicitsreasoning,
  title={Chain-of-thought prompting elicits reasoning in large language models},
  author={Wei, Jason and Wang, Xuezhi and Schuurmans, Dale and Bosma, Maarten and Xia, Fei and Chi, Ed and Le, Quoc V and Zhou, Denny and others},
  journal={Advances in neural information processing systems},
  volume={35},
  pages={24824--24837},
  year={2022}
}

@article{journals/debu/HamiltonYL17,
  author = {Hamilton, William L. and Ying, Rex and Leskovec, Jure},
  ee = {http://sites.computer.org/debull/A17sept/p52.pdf},
  journal = {IEEE Data Eng. Bull.},
  number = 3,
  pages = {52-74},
  title = {Representation Learning on Graphs: Methods and Applications.},
  url = {http://dblp.uni-trier.de/db/journals/debu/debu40.html#HamiltonYL17},
  volume = 40,
  year = 2017
}

@inproceedings{liu2019robertarobustlyoptimizedbert,
    title = "A Robustly Optimized {BERT} Pre-training Approach with Post-training",
    author = "Zhuang, Liu  and
      Wayne, Lin  and
      Ya, Shi  and
      Jun, Zhao",
    editor = "Li, Sheng  and
      Sun, Maosong  and
      Liu, Yang  and
      Wu, Hua  and
      Liu, Kang  and
      Che, Wanxiang  and
      He, Shizhu  and
      Rao, Gaoqi",
    booktitle = "Proceedings of the 20th Chinese National Conference on Computational Linguistics",
    month = aug,
    year = "2021",
    address = "Huhhot, China",
    publisher = "Chinese Information Processing Society of China",
    url = "https://aclanthology.org/2021.ccl-1.108/",
    pages = "1218--1227",
    language = "eng"
}

@inproceedings{robbani-etal-2024-flee,
    title = "Flee the Flaw: Annotating the Underlying Logic of Fallacious Arguments Through Templates and Slot-filling",
    author = "Robbani, Irfan  and
      Reisert, Paul  and
      Pothong, Surawat  and
      Inoue, Naoya  and
      Guerraoui, Cam{\'e}lia  and
      Wang, Wenzhi  and
      Naito, Shoichi  and
      Choi, Jungmin  and
      Inui, Kentaro",
    editor = "Al-Onaizan, Yaser  and
      Bansal, Mohit  and
      Chen, Yun-Nung",
    booktitle = "Proceedings of the 2024 Conference on Empirical Methods in Natural Language Processing",
    month = nov,
    year = "2024",
    address = "Miami, Florida, USA",
    publisher = "Association for Computational Linguistics",
    url = "https://aclanthology.org/2024.emnlp-main.1142/",
    doi = "10.18653/v1/2024.emnlp-main.1142",
    pages = "20524--20540"
}

@article{sourati2023robustexplainableidentificationlogical,
title = {Robust and explainable identification of logical fallacies in natural language arguments},
journal = {Knowledge-Based Systems},
volume = {266},
pages = {110418},
year = {2023},
issn = {0950-7051},
doi = {https://doi.org/10.1016/j.knosys.2023.110418},
url = {https://www.sciencedirect.com/science/article/pii/S0950705123001685},
author = {Zhivar Sourati and Vishnu Priya {Prasanna Venkatesh} and Darshan Deshpande and Himanshu Rawlani and Filip Ilievski and Hông-Ân Sandlin and Alain Mermoud}
}

@book{alma9924387453302466,
	address = {Cambridge},
	author = {Walton, Douglas N.},
	booktitle = {Informal logic : a pragmatic approach},
	edition = {Second edition.},
	isbn = {1-107-08657-4},
	language = {eng},
	publisher = {Cambridge University Press},
	title = {Informal logic : a pragmatic approach},
	year = {2008}}

@inproceedings{goffredo-etal-2023-argument,
    title = "Argument-based Detection and Classification of Fallacies in Political Debates",
    author = "Goffredo, Pierpaolo  and
      Chaves, Mariana  and
      Villata, Serena  and
      Cabrio, Elena",
    editor = "Bouamor, Houda  and
      Pino, Juan  and
      Bali, Kalika",
    booktitle = "Proceedings of the 2023 Conference on Empirical Methods in Natural Language Processing",
    month = dec,
    year = "2023",
    address = "Singapore",
    publisher = "Association for Computational Linguistics",
    url = "https://aclanthology.org/2023.emnlp-main.684/",
    doi = "10.18653/v1/2023.emnlp-main.684",
    pages = "11101--11112"
}

@inproceedings{yeh2024cocolofadatasetnewscomments,
    title = "{C}o{C}o{L}o{F}a: A Dataset of News Comments with Common Logical Fallacies Written by {LLM}-Assisted Crowds",
    author = "Yeh, Min-Hsuan  and
      Wan, Ruyuan  and
      Huang, Ting-Hao Kenneth",
    editor = "Al-Onaizan, Yaser  and
      Bansal, Mohit  and
      Chen, Yun-Nung",
    booktitle = "Proceedings of the 2024 Conference on Empirical Methods in Natural Language Processing",
    month = nov,
    year = "2024",
    address = "Miami, Florida, USA",
    publisher = "Association for Computational Linguistics",
    url = "https://aclanthology.org/2024.emnlp-main.39/",
    doi = "10.18653/v1/2024.emnlp-main.39",
    pages = "660--677"
}

@book{gensler2010z,
	author = {Gensler, H.J.},
	isbn = {9780810875968},
	number = {v. 169},
	publisher = {Bloomsbury Academic},
	series = {G - Reference,Information and Interdisciplinary Subjects Series},
	title = {The to Z of Logic},
	url = {https://books.google.it/books?id=-lrWH-aZw7QC},
	year = {2010}}

@misc{ruizdolz2025explainableframeworkmisinformationidentification,
      title={An Explainable Framework for Misinformation Identification via Critical Question Answering}, 
      author={Ramon Ruiz-Dolz and John Lawrence},
      year={2025},
      eprint={2503.14626},
      archivePrefix={arXiv},
      primaryClass={cs.CL},
      url={https://arxiv.org/abs/2503.14626}, 
}

@inproceedings{ruiz-dolz-lawrence-2023-detecting,
    title = "Detecting Argumentative Fallacies in the Wild: Problems and Limitations of Large Language Models",
    author = "Ruiz-Dolz, Ramon  and
      Lawrence, John",
    editor = "Alshomary, Milad  and
      Chen, Chung-Chi  and
      Muresan, Smaranda  and
      Park, Joonsuk  and
      Romberg, Julia",
    booktitle = "Proceedings of the 10th Workshop on Argument Mining",
    month = dec,
    year = "2023",
    address = "Singapore",
    publisher = "Association for Computational Linguistics",
    url = "https://aclanthology.org/2023.argmining-1.1/",
    doi = "10.18653/v1/2023.argmining-1.1",
    pages = "1--10"
}

@article{wangcabrio,
	author = {Xiaoou Wang and Elena Cabrio and Serena Villata},
	doi = {10.1177/19462174251330980},
	eprint = {https://doi.org/10.1177/19462174251330980},
	journal = {Argument \& Computation},
	number = {3},
	pages = {405-424},
	title = {When automated fact-checking meets argumentation: Unveiling fake news through argumentative evidence},
	url = {https://doi.org/10.1177/19462174251330980},
	volume = {16},
	year = {2025}}

@inproceedings{gutierrez2024detecting,
  author    = {Guti{\'{e}}rrez{-}Mandingorra, Ana and Heras, Stella and Palanca, Javier},
  title     = {Detecting Disinformation through Computational Argumentation Techniques and Large Language Models},
  booktitle = {Proceedings of the 24th Workshop on Computational Models of Natural Argument (CMNA 2024)},
  year      = {2024},
  series    = {CEUR Workshop Proceedings},
  volume    = {3769},
  pages     = {46--51},
  url       = {https://ceur-ws.org/Vol-3769/paper6.pdf}
}

\appendix

\section{Implementation Details}
In experiments where the task consisted of returning only the fallacy label, we set the temperature to 0, with the exception of \texttt{o4-mini}, \texttt{gpt-4o} and \texttt{deepseek-r1}. In all other experiments, the standard configuration was kept. Multiple prompt configurations were evaluated for each approach.


\section{Fallacy Datasets}
\label{sec:appendix-datasets}

\subsection{Logic}
The dataset \textsc{Logic} \citep{jin2022logicalfallacydetection} contains the following 13 fallacy classes: \textit{Faulty Generalization (Hasty Generalization)}, \textit{Ad Hominem}, \textit{Ad Populum}, \textit{Circular Claim (Circular Reasoning)}, \textit{False Cause (False Causality)}, \textit{Appeal to Emotion (Emotional Language)}, \textit{Fallacy of Relevance (Red Herring)}, \textit{Deductive Fallacy}, \textit{Intentional Fallacy}, \textit{Fallacy of Extension (Extension Fallacy)}, \textit{False Dilemma (Black-and-White Fallacy)}, \textit{Fallacy of Credibility (Irrelevant Authority)} and \textit{Equivocation}. The names in the parentheses are the actual names used in our experiments.

\subsection{Reddit}
The dataset \textsc{Reddit} \citep{sahai-etal-2021-breaking} contains 8 fallacy classes: \textit{Appeal to Authority (Irrelevant Authority)}, \textit{Appeal to Majority (Ad Populum)}, \textit{Appeal to Nature}, \textit{Appeal to Tradition}, \textit{Appeal to Worse Problems}, \textit{Black-and-White fallacy},  \textit{Hasty Generalization} and \textit{Slippery Slope}. It contains the class \textit{No Fallacy} as well. The names in parentheses are the actual labels used. In our experiments, only the classes included in \textsc{Logic} are retained (Table~\ref{tab:nonlogic-datasets}). We can keep the class \textit{Slippery Slope} because two generated patterns for \textit{Hasty Generalization} correspond to it. 

\subsection{ElecDebate}
The dataset \textsc{ElecDebate} \citep{goffredo-etal-2023-argument} contains the following 6 fallacy classes: \textit{Ad Hominem}, \textit{Appeal to Emotion (Emotional Language)}, \textit{Appeal to Authority (Irrelevant Authority)}, \textit{Slippery Slope}, \textit{False Cause} and \textit{Slogan}. The names in parentheses are the actual labels used. In our experiments, only the classes included in \textsc{Logic} are retained (Table~\ref{tab:nonlogic-datasets}). 


\begin{table}[t]
\centering
\setlength{\tabcolsep}{4pt}
\scalebox{0.75}{
\begin{tabular}{@{}p{0.55\columnwidth}p{0.45\columnwidth}@{}}
\toprule
\textbf{\textsc{Reddit}} & \textbf{\textsc{ElecDebate}} \\
\midrule
\begin{tabular}[t]{@{}l@{}}
    \textbullet~Ad Populum \\
    \textbullet~Irrelevant Authority \\
    \textbullet~Hasty Generalization \\
    \textbullet~Slippery Slope \\
    \textbullet~Black-and-White Fallacy
\end{tabular} &
\begin{tabular}[t]{@{}l@{}}
    \textbullet~Ad Hominem \\
    \textbullet~Irrelevant Authority \\
    \textbullet~Emotional Language \\
    \textbullet~Slippery Slope \\
    \textbullet~False Cause
\end{tabular} \\
\bottomrule
\end{tabular}
}
\caption{Fallacy classes in \textsc{Reddit} and \textsc{ElecDebate} used in our experiments.}
\label{tab:nonlogic-datasets}
\end{table}

\section{Baselines}
\label{sec:baseline}

We consider only the classes of \textsc{Reddit} and \textsc{ElecDebate} in common to \textsc{logic}. For this reason, direct comparison with prior work is generally not possible. However, for \textsc{reddit}, \citet{lei2024boostinglogicalfallacyreasoning} provide classwise F\textsubscript{1} scores, allowing us to compute Macro F\textsubscript{1} and compare our results. Tables~\ref{tab:reddit-baseline} and \ref{tab:elecdebate-baseline} present the comparison with prior work for both datasets.

\begin{table}[h!]
\centering
\renewcommand{\arraystretch}{1.1}
\setlength{\tabcolsep}{3pt}
\scalebox{0.85}{
\begin{tabular}{p{4cm}c}
\toprule
\textbf{Method} & \textbf{Macro F\textsubscript{1}} \\
\midrule
\multicolumn{2}{l}{\emph{Supervised}} \\
\quad \citet{sahai-etal-2021-breaking} & 58.4 \\
\quad \citet{lei2024boostinglogicalfallacyreasoning}\textsuperscript{\dag} & 81.3 \\
\quad \citet{pan2024llmsgoodzeroshotfallacy} & 83.2 \\
\multicolumn{2}{l}{\emph{Unsupervised}} \\
\quad \citet{pan2024llmsgoodzeroshotfallacy} & 81.1 \\
\quad \citet{yeh2024cocolofadatasetnewscomments} & 81.0 \\
\multicolumn{2}{l}{\emph{Ours}} \\
\quad \textbf{\textsc{patterns}} & \textbf{84.5} \\
\quad \textbf{\textsc{same-dataset pattern matching}} & \textbf{84.3} \\
\bottomrule
\end{tabular}
}
\caption{Performance comparison on \textsc{Reddit}.\textsuperscript{\dag} indicates that Macro F\textsubscript{1} is computed on the exact same classes as \textsc{logic}.}
\label{tab:reddit-baseline}
\end{table}

\begin{table}[h!]
\centering
\renewcommand{\arraystretch}{1.1}
\setlength{\tabcolsep}{3pt}
\scalebox{0.85}{
\begin{tabular}{p{4cm}c}
\toprule
\textbf{Method} & \textbf{Macro F\textsubscript{1}} \\
\midrule
\multicolumn{2}{l}{\emph{Supervised}} \\
\quad \textbf{\citet{goffredo-etal-2023-argument}} & \textbf{73.9} \\
\quad \citet{pan2024llmsgoodzeroshotfallacy} & 62.3 \\
\multicolumn{2}{l}{\emph{Unsupervised}} \\
\quad \citet{pan2024llmsgoodzeroshotfallacy} & 44.5 \\
\multicolumn{2}{l}{\emph{\textbf{Ours}}} \\
\quad \textsc{same-dataset pattern} & 64.9 \\
\quad \textsc{dynamic one-shot} & 70.4 \\
\bottomrule
\end{tabular}
}
\caption{Performance comparison on \textsc{elecdebate}.}
\label{tab:elecdebate-baseline}
\end{table}

\section{Additional Experiments}
\label{sec:additiona-experiments}
We are going to report some other experimental set-ups that have been explored, including some basic baselines that we have not included in Section~\ref{sec:experiments}.

\subsection{Prompt design}
\begin{itemize}

    \item \textbf{\textsc{exp}}: to investigate whether explicit reasoning improves performance, we implemented a baseline that not only provides fallacy names but also requests the model to generate a two-sentence explanation for its classification decision,
    testing whether forcing the model to articulate its reasoning leads to better outcomes. 
    The two-sentence constraint was intentionally designed to keep explanations concise and manageable for manual inspection of explanations.

    \item \textbf{\textsc{guidelines}}: to leverage the model's classification errors for improvement, we develop guidelines derived from observed mistakes. We conduct pattern matching evaluation on the validation set
    and collect misclassified instances. For each class, we provide the model with incorrectly classified examples and prompt it to generate comprehensive detection guidelines (as can be seen from table~\ref{tab:irr-guideline}), given our generated pattern as a reference. 
    These guidelines are then adopted to evaluate the test set. Notably, 
    only guidelines produced by \texttt{o4-mini} and partially by \texttt{gpt-4.1-mini} incorporate a little structural and logical information such as common connectors or logical forms while the majority of guidelines content across models focuses primarily on semantic characteristics rather than structural patterns.


\end{itemize}


\begin{table}[t]
\renewcommand{\arraystretch}{1.05}
\centering
\fontsize{10.0pt}{10.5pt}\selectfont
\begin{tabular}{@{}p{0.16\textwidth} p{0.30\textwidth}@{}}
\toprule
\textbf{Fallacy} & \textbf{Irrelevant Authority} \\
\midrule
\textbf{Core definition} &
A fallacy that treats an individual’s status, title, or popularity as proof of a claim when their expertise or relevance to the topic is absent or insufficient. \\
\midrule
\textbf{Key indicators} &
Argument rests on ``X says so'' without independent support.\newline
Authority cited has no recognized expertise in the claim’s domain.\newline
No substantive evidence beyond the authority’s endorsement. \\
\midrule
\textbf{Typical confusion patterns} &
Ad Populum: group popularity vs. single authority endorsement.\newline
Appeal to Tradition: ``has always been done by experts'' vs. citing irrelevant experts.\newline
Equivocation: shifting word senses vs. relying on irrelevant credentials. \\
\bottomrule
\end{tabular}
\caption{Guidelines relative to the \textit{Irrelevant Authority} fallacy generated by \texttt{o4-mini}.}
\label{tab:irr-guideline}
\end{table}

\subsection{Results}


\begin{table}[!t]
\centering
\fontsize{8.7pt}{9.8pt}\selectfont
\setlength{\tabcolsep}{4pt}
\renewcommand{\arraystretch}{1.08}
\begin{tabular}{@{}lcccc@{}}
\toprule
\textbf{Model} 
& \multicolumn{2}{c}{\textbf{\textsc{exp}}}
& \multicolumn{2}{c}{\textbf{\textsc{guidelines}}} \\
\cmidrule(lr){2-3} \cmidrule(lr){4-5}
& \textbf{Acc.} & \textbf{F\textsubscript{1}}
& \textbf{Acc.} & \textbf{F\textsubscript{1}} \\
\midrule
\texttt{o4-mini}        & 61.3 & 61.5 & 65.5 & 65.7 \\
\texttt{gpt-4o}         & 60.4 & 53.3 & 62.6 & 56.6 \\
\texttt{deepseek-r1}    & 61.8 & 54.8 & 63.5 & 57.9 \\
\texttt{gpt-4.1-mini}   & 57.5 & 57.9 & 60.5 & 60.6 \\
\texttt{llama-3.3-70B}  & 56.1 & 56.5 & 52.8 & 53.3 \\
\texttt{gemma-3-27b-it} & 59.1 & 60.8 & 58.8 & 59.4 \\
\bottomrule
\end{tabular}
\caption{Logical fallacy classification performance on additional experiments. F\textsubscript{1} denotes Macro-F\textsubscript{1}.}
\label{tab:logic-onecol}
\end{table}

\textbf{\textsc{exp}}'s (Table~\ref{tab:logic-onecol}) results show that requesting the model to articulate the reasoning does not really cause any improvement. Specifically, certain classes such as \textit{Intentional Fallacy} and \textit{Extension Fallacy} exhibit extremely low F\textsubscript{1} scores under the non-reasoning models (0.027 and 0.13 respectively on average), indicating performance deterioration compared to the \textbf{\textsc{zero-shot}} baseline. This proves that models process surface-level semantic patterns without being able to access the multi-layered intentional structures behind reasoning (Table~\ref{tab:wrong-patterns}). 

\begin{table}[!t]
\renewcommand{\arraystretch}{1.08}
\centering
\fontsize{9.3pt}{10.5pt}\selectfont
\begin{tabular}{@{}p{0.31\columnwidth} p{0.47\columnwidth} p{0.18\columnwidth}@{}}
\toprule
\textbf{Text} & \textbf{Explanation} & \textbf{Gold} \\
\midrule
The Bible is true because God exists, and God exists because the Bible says so.
&
The argument uses its conclusion as a premise, creating a logical loop without independent evidence.
\newline
\textbf{\emph{Circular Reasoning}}
&
\textbf{\emph{Circular Reasoning}} \\
\midrule
My friend said that if you sneeze more than three times, you have the corona virus.
&
The argument assumes sneezing three times indicates the virus, generalizing a symptom without considering other causes.
\newline
\textbf{\emph{Hasty Generalization}}
&
\textbf{\emph{Irrelevant Authority}} \\
\bottomrule
\end{tabular}
\caption{Examples from \texttt{GPT-4.1-mini} in the \textbf{\textsc{exp}} setting: the first is correctly classified; the second is misclassified because the explanation, while coherent, fails to capture the underlying fallacy.}
\label{tab:wrong-patterns}
\end{table}

Including \textbf{\textsc{guidelines}} yields only modest results. While these guidelines are designed to provide comprehensive fallacy knowledge, they appear to lack the appropriate type of information from which models can benefit. Indeed, providing explicit information about the underlying logical structure proves significantly more beneficial for model performance.


\section{Syntax-augmented roBERTa}
\label{sec:syntax-augm}
\citet{sachan2021syntaxtreeshelppretrained} introduces a syntax-augmented model that incorporates dependency tree information into pre-trained BERT-based \citep{devlin-etal-2019-bert} transformers through specialized Graph Neural Networks (GNNs) \citep{journals/debu/HamiltonYL17} that process dependency trees. The authors introduce two distinct fusion strategies to integrate syntactic structure into BERT representation. We adopted specifically roBERTa-large \citep{liu2019robertarobustlyoptimizedbert} in the attempt to perform a syntax-driven examples selection. Further details about the implementation are available in \citet{sachan2021syntaxtreeshelppretrained}.

\end{document}